\documentclass{article}
\usepackage{multirow,array}
\usepackage{spconf,amsmath,graphicx,tabu}
\usepackage{mathrsfs}
\usepackage{amssymb,booktabs}

\title{driving maneuvers prediction based on cognition-driven and data-driven method}
\name{Dong Zhou \qquad Huimin Ma \qquad Yuhan Dong\thanks{This work was supported by the National Key R\&D Plan  (No. 2016YFB0100901) and the National Natural Science Foundation of China  (No. 61171113 and No. 61773231).}}
\address{Department of Electronic Engineering, Tsinghua University, Beijing 100084, China\\
zhoud16@mails.tsinghua.edu.cn, mhmpub@tsinghua.edu.cn, dongyuhan@sz.tsinghua.edu.cn}
\begin{document}
%
\maketitle

\begin{abstract}
Advanced Driver Assistance Systems (ADAS) improve driving safety significantly. They alert drivers from unsafe traffic conditions when a dangerous maneuver appears. Traditional methods to predict driving maneuvers are mostly based on data-driven models alone. However, existing methods to understand the driver's intention remain an ongoing challenge due to a lack of intersection of human cognition and data analysis. To overcome this challenge, we propose a novel method that combines both the cognition-driven model and the data-driven model. We introduce a model named Cognitive Fusion-RNN (CF-RNN) which fuses the data inside the vehicle and the data outside the vehicle in a cognitive way. The CF-RNN model consists of two Long Short-Term Memory  (LSTM) branches regulated by human reaction time. Experiments on the Brain4Cars benchmark dataset demonstrate that the proposed method outperforms previous methods and achieves state-of-the-art performance.
\end{abstract}
\begin{keywords}
Image cognition, data fusion, ADAS, Recurrent neural networks, driving maneuvers prediction, CF-RNN
\end{keywords}
\section{Introduction}
\label{sec:intro}

Many people die in traffic accidents every year. In the US, more than 35,000 people died in road accidents in 2015, the majority of which were caused by improper driving maneuvers on motor vehicles \cite{ref1}. Some applications in Advanced Driver Assistance Systems (ADAS) have been proposed to alert drivers from dangerous maneuvers. Jain \emph{et al}.  \cite{Jain2015, Jain2016} predicted driving maneuvers by data fusion from the collected videos, Global Positioning System (GPS), and other outside information. Zyner \emph{et al}. \cite{Zyner2017} used the position, GPS, Inertial Measurement Unit  (IMU) and odometry data to understand the driver's intention. Ortiz \emph{et al}. \cite{Ortiz2011} used vehicle speed measurements from the Controller Area Network (CAN) bus, along with the traffic light sensing data for predicting driver braking behavior. In general, the collected data in the driving maneuvers prediction task could be divided into two classes using driver's perception as a boundary. One is the environment information outside the vehicle (e.g., GPS, IMU, CAN bus data), the other is the driver's performance inside the vehicle (e.g., eye gaze, head pose,  movement of face key points). However, predicting the future driving maneuvers is not only a data-driven task but also an important cognition-driven task. Relying on the data-driven model alone cannot truly reflect the human cognition process. To make the prediction model more intelligent and accurate, we propose a model which combines the data-driven model with the cognition-driven model. A cognition-driven model means to model driving behavior from the cognitive perspective. The cognition-driven model in Fig. 1 shows the relationships between driver's perception-reaction time and the status change of vehicle and environment. After perceiving and cognizing the outside information, the driver makes a decision related to the current driving environment, which finally reflects in the driving maneuvers.

\begin{figure}[t]
\begin{minipage}[b]{1.0\linewidth}
  \centering
  \centerline{\includegraphics[width=9.0cm]{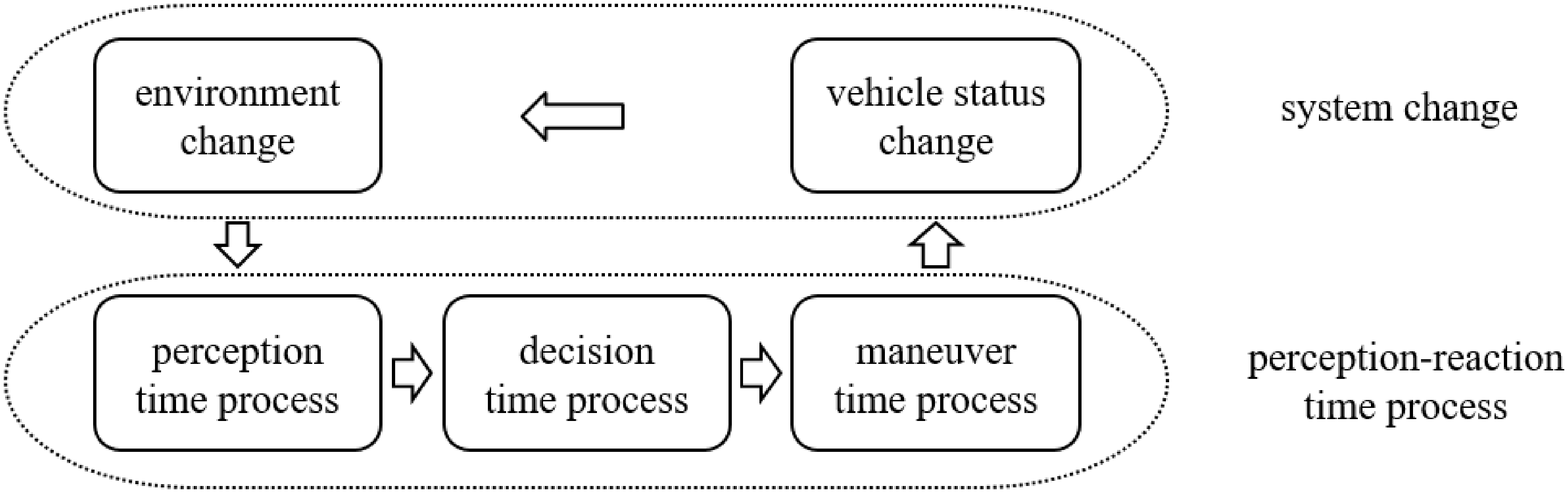}}
\end{minipage}
\caption{Cognition-driven model to represent the relationships between driver's perception-reaction time process and the status change of vehicle and environment}
\label{fig:res}
\end{figure}

\begin{figure*}[htb]

\begin{minipage}[b]{1.0\linewidth}
  \centering
   \centerline{\includegraphics[width=17.5cm, height=8.0cm]{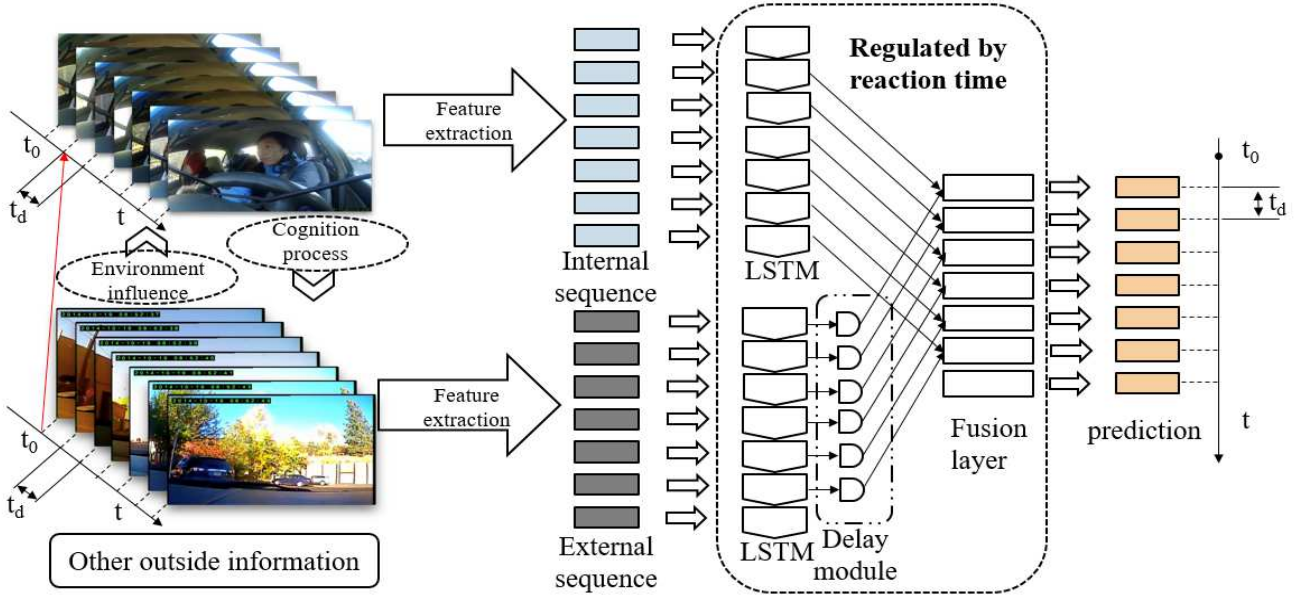}}

\end{minipage}
\caption{The brief description of the proposed Cognitive Fusion-RNN structure}
\label{fig:res}
\end{figure*}

Some previous works attempted to build the bridge between human cognition and the outside driving environment. An integrated driver model \cite{Salvucci2006} was proposed from the Adaptive Control of Thought-Rational (ACT-R) cognitive architecture. Addario \emph{et al}. \cite{Addario2014} studied the driver's ability for detection and response to emergency roadway hazards by modeling a cognitive structure. Wickens \emph{et al.} proposed an adaption model \cite{Wickens2000} of environment information processing concerning the stages of driver's perception-response time. In the classic optimal velocity (OV) model \cite{MBandoKHasebeaNakayamaaShibata1995}, the driver's reaction time appears as an important parameter. These studies show that driving actions are obviously influenced by the outside environment and the state of the vehicle. But these methods are theoretical or classic so that they lack the combination of the latest data-driven approaches and the cognition theory. To solve the problem, our model combines the latest data analysis method with the cognition process.

The main contribution of this paper is the establishment of Cognitive Fusion Recurrent Neural Networks (CF-RNN) model based on cognition-driven model and data-driven model. The CF-RNN consists of two Long Short-Term Memory (LSTM) units from both inside and outside of the car. The outputs of the two LSTM units are regulated by human cognition time process. Besides that, we also provide the necessary and adequate perception and cognition time of the drivers in the Brain4Cars dataset \cite{Jain2015} to estimate the average reaction time and generate best prediction results. This paper is organized as follows. Sec. 2 introduces the details of the proposed CF-RNN method. Sec. 3 presents the experiments and analysis, and the conclusion is made in Sec. 4.

\section{The proposed method}
\label{sec:format}

\subsection{Traditional fusion model}
\label{ssec:subhead}

From the data-driven perspective, the driving maneuvers prediction problem could be modeled as a temporal sequence learning problem. When we process the driving maneuvers prediction task by multi-source data fusion, LSTM\cite{Hochreiter1997} networks are widely used. In many applications about driver's behavior, LSTM performs better than traditional model and standard RNN \cite{Jain2016, Zyner2017, Morton2017, Wollmer2011, Olabiyi, Donahue2017, Yue-HeiNgJoeandHausknechtMatthewandVijayanarasimhanSudheendraandVinyalsOriolandMongaRajatandToderici2015}. Some previous works analyze the importance and superiority of LSTM for modeling of driver's behavior \cite{Morton2017}. In the driver's operation scenario, the driver processes the environment information just based on the few seconds before current moment. So it is a wise choice to use LSTM or other variety of LSTM to solve the sequence learning problem. Our objective is to achieve the combination of the cognition-driven model and the data-driven model. As already introduced, the data could be divided into two classes: external data and internal data. Without any inspiration from the real driving situation, the traditional methods establish the fusion model, which fuses the LSTM output high-level representations from both inside and outside of the vehicle at the same time  \cite{Jain2016}. To fuse the outside data and inside data like a human in an actual driving scene, we propose a method which prioritizes external information and combines the two different sources of data cognitively.

\subsection{Cognitive fusion model}
\label{ssec:subhead}

To model the video sequence learning problem, two independent LSTM units  (one for outside data and the other for data inside the car) are used. Our method is inspired by an intuitive idea that external features and internal features should not be fused at the same time, caused by the time consumption of human reaction. There is an objectively existence of time delay due to human perception and cognition between external variables and internal variables. The time duration $t_{d}$ represents the time between driver's perception of the external change and actions. From a temporal perspective, the variables of real driving scenarios are continuous, but the video sequence learning problem is discrete. We could sample from the data within the time interval $t_{d}$ because an action needs $t_{d}$ time duration before it happens. The observations of external variables over $T$ time steps form a sequence of vectors $\boldsymbol{x_{t_{0}}}$, $\boldsymbol{x_{t_{0}+t_{d}}}$, $\boldsymbol{x_{t_{0}+2t_{d}}}$, \ldots, $\boldsymbol{x_{t_{0}+ (T-1)t_{d}}}$, and the observations of internal variables over $T$ time steps form a sequence of vectors $\boldsymbol{z_{t_{0}}}$, $\boldsymbol{z_{t_{0}+t_{d}}}$, $\boldsymbol{z_{t_{0}+2t_{d}}}$, \ldots, $\boldsymbol{z_{t_{0}+ (T-1)t_{d}}}$. And the current LSTM cell memory states at each time step are denoted as $\boldsymbol{M_{t_{0}}}$, $\boldsymbol{M_{t_{0}+t_{d}}}$, $\boldsymbol{M_{t_{0}+2t_{d}}}$, \ldots, $\boldsymbol{M_{t_{0}+ (T-1)t_{d}}}$. High level representations of the current LSTM cell at each time step are denoted as $\boldsymbol{H_{t_{0}}}$, $\boldsymbol{H_{t_{0}+t_{d}}}$, $\boldsymbol{H_{t_{0}+2t_{d}}}$, \ldots, $\boldsymbol{H_{t_{0}+ (T-1)t_{d}}}$. The final prediction after fusion are denoted as $\boldsymbol{y_{t_{0}}}$, $\boldsymbol{y_{t_{0}+t_{d}}}$, $\boldsymbol{y_{t_{0}+2t_{d}}}$, \ldots, $\boldsymbol{y_{t_{0}+ (T-1)t_{d}}}$. We use $y^{k}_t$ to represent the probability of the temporal sequence belonging to the $k$ event  (total number of maneuver events is $K$) at the $t$ point of time. So it is obvious that $\sum\limits_{k=1}^{K} y^{k}_{t} = 1$. Each $\boldsymbol{y_{t}}$ is a one-hot vector in which only one event in $K$ should be encoded as probability $1$. The symbol $t$ which uses $t_{d}$ as interval, represents the sequence index from $t_{0}$ to $t_{0}+ (T-1)t_{d}$. And we denote the operation of the LSTM unit as function $L$ ($L_{x}$ for the sequence of $\boldsymbol{x_{t}}$ and $L_{z}$ for the sequence of $\boldsymbol{z_{t}}$) and the operation of the fusion layer with softmax layer as $F$. Based on the cognition theory, $\boldsymbol{x_{t}}$ and $\boldsymbol{z_{t}}$ should have a relative delay which we denote as $t_{d}$. So the prediction formula for our proposed Cognitive Fusion-RNN could be written as:

\begin{align}
& \boldsymbol{ (H^x_{t}, M^x_{t})} = L_{x}\boldsymbol{ (x_{t}, M^x_{t-t_{d}}, H^x_{t-t_{d}})} \\
& \boldsymbol{ (H^z_{t+t_{d}}, M^z_{t+t_{d}})} = L_{z}\boldsymbol{ (z_{t+t_{d}}, M^z_{t}, H^z_{t})}  \label{q} \\
& \boldsymbol{y_{t+t_{d}}} = F (\boldsymbol{H^x_{t}}, \boldsymbol{H^z_{t+t_{d}})}
\end{align}

Fig. 2 is the brief description of the proposed CF-RNN structure. The different branch of LSTM learns the different temporal features from the different source, so we need to introduce the control of the reaction time for the data fusion. In the CF-RNN model, we add a time delay module before the high-level features fusion. Information at the beginning and the end should be specially processed because it could not find the corresponding data to match. There are two methods called ``padding'' method and ``margin'' method to match the internal sequence and external sequence. As shown in Fig. 3, we could find that ``padding'' method remains all the information but the coupling at the beginning and the coupling at the end of the sequence are not so strong. The ``margin'' method loses some information, but it increases the coupling of the two sequences.

\begin{figure}[htb]
\begin{minipage}[b]{1.0\linewidth}
  \centering
  \centerline{\includegraphics[width=9.0cm]{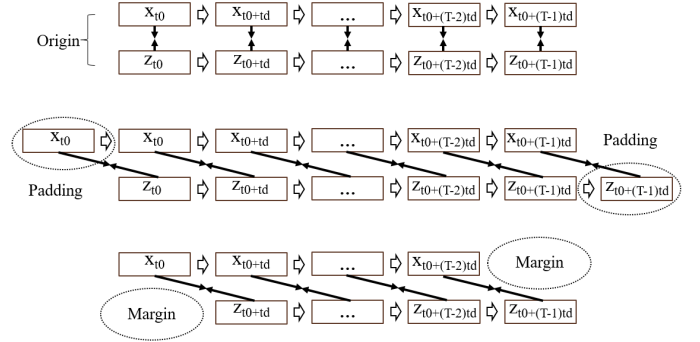}}
\end{minipage}
\caption{Structures in the figure from top to bottom are three data fusion methods : origin fusion method, ``padding'' method and ``margin'' method.}
\label{fig:res}
\end{figure}

The driving maneuvers prediction task has a strong real-time property which needs ADAS alert the driver in a few seconds before the driver making mistakes. So in the CF-RNN architecture, time of driver's perception and reaction which denoted as $t_{d}$ should be chosen cautiously. Little change of $t_{d}$ could make a difference. Driver's reaction time is significantly affected by many factors such as driver's age, gender, fatigue, distraction and intelligence \cite{Kosinski2013}. The time delay $t_{d}$ is also highly variable. For different drivers, the $t_{d}$ could be different. However, we just need an approximate estimate of $t_{d}$. One popular opinion which stems from Olson and Sivak \cite{Olson1986Perception} is that $t_{d}$ equals to 1.10 seconds. In the dynamic control model of vehicles \cite{MBandoKHasebeaNakayamaaShibata1995}, the time delay $t_{d}$ is equal to about 1.00 seconds. In the next section, we explore the best chosen of $t_{d}$ on the benchmark.

\begin{table*}[htbp]
\caption{Evaluation of the Effectiveness of Our Methods. Standard error is also shown. Algorithms are compared on the features  (features of head pose are 2D) from \cite{Jain2015}.}
\small
\scalebox{1.02}{
\begin{tabular}{lcccp{1.5cm}<{\centering}cccp{1.5cm}<{\centering}}
\toprule
\multirow{2}{*}{Method} &
\multicolumn{4}{c}{Lane change} &
\multicolumn{4}{c}{Turns}\\
\cline{2-9}   
& Pr (\%) & Re (\%) & F1-s (\%) & Time-to-maneuvers(s)& Pr (\%) & Re (\%) & F1-s (\%) & Time-to-maneuvers(s) \\
\midrule
FRNN-UL & $\boldsymbol{92.7} \pm 2.1$ & $84.4 \pm 2.8$ & $88.3$ & $3.46$ & $81.2 \pm 3.5
$ & $78.6 \pm 2.8$ & $79.9$ & $3.94$\\
FRNN-EL & $88.2 \pm 1.4$ & $86.0 \pm 0.7$ & $87.1$ & $3.42$ & $83.8 \pm 2.1$ & $79.9 \pm 3.5$ & $81.8$ & $3.78$\\
CF-RNN (padding) & $87.4 \pm 1.7$ & $91.8 \pm 1.6$ & $89.5$ &$3.35$ & $84.7 \pm 3.8$ & $81.1 \pm 3.1$ & $82.8$ & $3.10$\\
CF-RNN (margin) & $87.3 \pm 1.0$ & $\boldsymbol{93.8} \pm 1.7$ & $\boldsymbol{90.5}$ & $3.75$ & $\boldsymbol{86.0} \pm 2.1$ & $\boldsymbol{81.4} \pm 1.7$ & $\boldsymbol{83.6}$ & $3.44$\\
\bottomrule
\end{tabular}}
\end{table*}

\section{EXPERIMENTS}
\label{sec:pagestyle}

\subsection{Brain4Cars Dataset}
\label{ssec:subhead}

The dataset used as a benchmark for our driving maneuvers prediction task is publicly released by the Brain4Cars team \cite{Jain2015}. This dataset consists of 5 classes of total 700 maneuvers data: left lane change, right lane change, left turn, right turn and driving straight. The problem could be divided into three subproblems: lane change, turns, and all maneuvers. Each maneuver includes a pair of videos with a duration of about 5 seconds: one for the driver's face variables inside the car, and the other one for the outside road environment. Additional information data are provided for each frame, including lane configuration, the presence of intersections ahead of the car and the speed.

\begin{table}[htbp]
\caption{Algorithms are compared on the features (features of head pose are 3D) from \cite{Jain2016}.}
\small
\scalebox{0.77}{\begin{tabular}{lcccp{1.5cm}<{\centering}}
\toprule
\multirow{2}{*}{Method} &
\multicolumn{4}{c}{All maneuvers}\\
\cline{2-5}   
& Pr (\%) & Re (\%) & F1-s (\%) & Time-to-maneuvers(s) \\
\midrule
FRNN-EL w/ 3D head pose & $90.5 \pm 1.0 $ & $87.4 \pm 0.5 $ & $88.9$ & $3.16$ \\
CF-RNN (padding) & $89.7 \pm 2.3$ & $89.4 \pm 1.7$ & $89.5$ & $3.03$\\
CF-RNN (margin) & $\boldsymbol{91.7} \pm 2.2$ & $\boldsymbol{90.7} \pm 2.4$ & $\boldsymbol{91.2}$ & $3.30$\\
\bottomrule
\end{tabular}}
\end{table}

\subsection{Evaluation setup}
\label{ssec:subhead}

To compare our model with the previous models, we use the same feature extraction pipeline as Jain \emph{et al}. \cite{Jain2016} with some modifications to sampling interval.  We change the dimension of the features due to the changed sampling interval. The videos are 25 frames per second so that the duration time of one frame is 0.04 seconds. At each time step t, $\boldsymbol{x_{t}}$ and $\boldsymbol{z_{t}}$ are computed over the last 0.80 seconds  (20 frames) of driving information \cite{Jain2016}. So the time interval between two adjacent input sampled frames $\boldsymbol{x_{t}}$ and $\boldsymbol{x_{t+t_{d}}}$ is 0.80 seconds (20 frames). Then we put $\boldsymbol{x_{t-t_{d}}}$ and $\boldsymbol{z_{t}}$ into the different LSTM units for training. After training of LSTM, we put the high-level output of the different LSTM unit together into the fusion layer. For the evaluation of our method, we train each model 10 times and calculate the average results. We evaluate the precision  ($P_{r}$) and recall  ($R_{e}$) which are defined as in \cite{Jain2015} for an anticipation algorithm. And to balance the precision and recall, we use F1\textrm{-}score to evaluate the performance of different models: $F1\textrm{-}score = \frac{2 \times P_{r} \times R_{e}}{ P_{r} + R_{e}}$. We also calculate the time to maneuvers which presents the time duration between the time of algorithm’s prediction and the start of the maneuver. Network training runs on a machine with GPU NVIDIA TITAN X. About half an hour on the machine is enough for training one time.

\subsection{Results}
\label{ssec:subhead}

Two fusion methods shown in Fig. 3 are optional. Table 1 and Table 2 show the results of our comparison with previous models. We can find that both of our algorithms can significantly improve the recall. And in the prediction task of turns and all maneuvers, both precision and recall are increased. The results show that even though the ``margin'' method lose some information at the beginning and information at the end, it still performs better than ``padding'' due to better coupling of sequences. The results demonstrate that the cognition time process is a critical time process in driving, and we should combine the cognition-driven method with the data-driven method.

To explore the best model for driving maneuvers prediction, we change the value of perception and cognition time $t_{d}$ from 0.52 seconds to 1.00 seconds and retrain the CF-RNN model. We choose 0.04 seconds for the change interval.  Fig. 4 describes the results of the experiments. From Fig. 4 we could find that F1\textrm{-}score performs best at $t_{d}$ = 0.84 seconds. We finally get the best average F1\textrm{-}score = \textbf{92.1}\%. At the same time, the average precision = \textbf{92.0}\% and average recall = \textbf{92.3}\%. The results demonstrate that the average perception and cognition time of the drivers in the dataset is about 0.84 seconds. After considering the cognition time process, the performance increases gradually, which shows that the combination of the cognition-driven method and the data-driven-method is effective.

\begin{figure}[htb]

\begin{minipage}[b]{1.0\linewidth}
  \centering
  \centerline{\includegraphics[width=8.5cm]{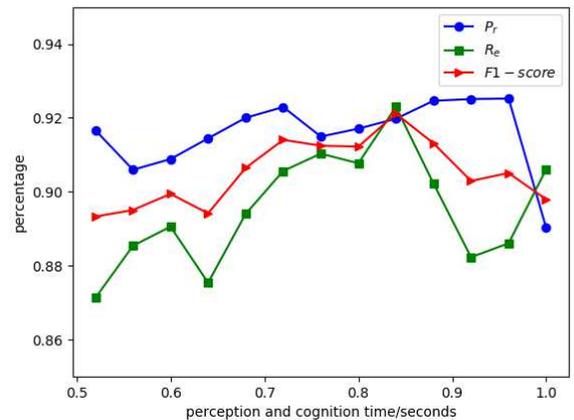}}
\end{minipage}
\caption{The relationship between $t_{d}$ and $P_{r}$, $R_{e}$ and $F1\textrm{-}score$.}
\label{fig:res}
\end{figure}

\section{CONCLUSION}

In this work, we considered the problem of driving maneuvers prediction in ADAS. The driving process usually consists of three subprocesses: perception, cognition, and action. Inspired by the driver's cognition and reaction system, we introduce the time delay module in the traditional data fusion structure to make the algorithm more like a real human in driving task. More importantly, we take full account of the cognition process and propose a novel method which combines the cognition-driven method with the data-driven method for the driving maneuvers prediction task. Our Cognitive Fusion-RNN model achieves state-of-art performance by improving F1\textrm{-}score from 88.9\% to 92.1\% on Brain4Cars dataset.

\label{sec:typestyle}
\bibliographystyle{IEEEbib}
\bibliography{strings,refs}

\end{document}